\documentclass[runningheads]{llncs}
\usepackage{graphicx}

\usepackage{tikz}
\usepackage{comment}
\usepackage{amsmath,amssymb} 
\usepackage{mathrsfs}  
\usepackage{color}
\usepackage{dsfont}
\usepackage{wrapfig}

\usepackage{xcolor,colortbl} 
\definecolor{bestcol}{RGB}{99, 255, 105}
\definecolor{secondbestcol}{RGB}{250, 255, 99}
\definecolor{worstcol}{RGB}{255, 141, 99}
\usepackage{arydshln}
\usepackage{subcaption}

\usepackage[accsupp]{axessibility}  

\begin{document}
\pagestyle{headings}
\mainmatter
\def\ECCVSubNumber{100}  

\title{How Well Do Vision Transformers (VTs) Transfer To The Non-Natural Image Domain? An Empirical Study Involving Art Classification}

\titlerunning{How Well Do VTs Transfer To The Non-Natural Image Domain?}
%
\author{Vincent Tonkes\inst{1} \and
Matthia Sabatelli\inst{1}}
\authorrunning{V. Tonkes \& M. Sabatelli}
%
\institute{Department of Artificial Intelligence and
Cognitive Engineering, University of Groningen, 9712 CP Groningen, The
Netherlands \\
\email{\{m.sabatelli\}@rug.nl}}
\maketitle

\begin{abstract}
Vision Transformers (VTs) are becoming a valuable alternative to Convolutional Neural Networks (CNNs) when it comes to problems involving high-dimensional and spatially organized inputs such as images. However, their Transfer Learning (TL) properties are not yet well studied, and it is not fully known whether these neural architectures can transfer across different domains as well as CNNs. In this paper we study whether VTs that are pre-trained on the popular ImageNet dataset learn representations that are transferable to the non-natural image domain. To do so we consider three well-studied art classification problems and use them as a surrogate for studying the TL potential of four popular VTs. Their performance is extensively compared against that of four common CNNs across several TL experiments. Our results show that VTs exhibit strong generalization properties and that these networks are more powerful feature extractors than CNNs.  
\end{abstract}

\section{Introduction}
\label{sec:introduction}
Since the introduction of AlexNet, roughly a decade ago, Convolutional Neural Networks (CNNs) have played a significant role in Computer Vision (CV) \cite{krizhevsky2012imagened}. Such neural networks are particularly well-tailored for vision-related tasks, given that they incorporate several inductive biases that help them deal with high dimensional, rich input representations. As a result, CNNs have found applications across a large variety of domains that are not \textit{per-se} restricted to the realm of natural images. Among such domains, the Digital Humanities (DH) field is of particular interest. Thanks to a long tradition of works that aimed to integrate advances stemming from technical disciplines into the Humanities, they have been serving as a challenging real-world test-bed regarding the applicability of CV algorithms. It naturally follows that over the last years, several works have studied the potential of CNNs within the DH (see \cite{fiorucci2020machine} for a survey about the topic), resulting in a significant number of successful applications that range from the classification of artworks \cite{tan2016ceci,sabatelli2018deep,zhong2020fine,milani2021dataset} to the detection of objects within paintings \cite{gonthier2018weakly,sabatelli2021advances}, automatic style classification \cite{chu2018image} and even art understanding \cite{bai2021explain}.

A major breakthrough within the CV community has recently been achieved by the Vision Transformer (VT) \cite{dosovitskiy2020image}, a novel neural architecture that has gained state-of-the-art performance on many standard learning benchmarks including the popular ImageNet dataset \cite{deng2009imagenet}. Exciting as this may be, we believe that a plain VT is not likely to become as valuable and powerful as a CNN as long as it does not exhibit the strong generalization properties that, over the years, have allowed CNNs to be applied across almost all domains of science \cite{van2014transfer,Mormont_2018_CVPR_Workshops,ackermann2018using}. Therefore, this paper investigates what VTs offer within the DH by studying this family of neural networks from a Transfer Learning (TL) perspective. Building on top of the significant efforts that the DH have been putting into digitizing artistic collections from all over the world \cite{strezoski2017omniart}, we define a set of art classification problems that allow us to study whether pre-trained VTs can be used outside the domain of natural images. We compare their performance to that of CNNs, which are well-known to perform well in this setting, and present to the best of our knowledge the very first thorough empirical analysis that describes the performance of VTs outside the domain of natural images and within the domain of art specifically.

\section{Preliminaries}
\label{sec:preliminaries}
We start by introducing some preliminary background that will be used throughout the rest of this work. We give an introduction about supervised learning and transfer learning (Sec. \ref{sec:tl}), and then move towards presenting some works that have studied Convolutional Neural Networks and Vision Transformers from a transfer learning perspective. 

\subsection{Transfer Learning}
\label{sec:tl}

With Transfer Learning (TL), we typically denote the ability that machine learning models have to retain and reuse already learned knowledge when facing new, possibly related tasks \cite{pan2009survey,zhuang2020comprehensive}. While TL can present itself within the entire machine learning realm \cite{bengio2012deep,taylor2009transfer,zhu2020transfer,ying2018transfer}, in this paper we consider the supervised learning setting only, a learning paradigm that is typically defined by an input space $\mathcal{X}$, an output space $\mathcal{Y}$, a joint probability distribution $P(X,Y)$ and a loss function $\ell: \mathcal{Y} \times \mathcal{Y} \rightarrow \mathds{R}$. The goal of a supervised learning algorithm is to learn a function $f:\mathcal{X}\rightarrow\mathcal{Y}$ that minimizes the expectation over $P(X,Y)$ of $\ell$ known as the expected risk. Classically, the only information that is available for minimizing the expected risk is a learning set $\mathcal{L}$ that provides the learning algorithm with $N$ pairs of input vectors and output values $(\mathbf{x}_1, y_1),...,(\mathbf{x}_N, y_N)$ where $\mathbf{x}_i \in \mathcal{X}$ and $y_i \in \mathcal{Y}$ are i.i.d. drawn from $P(X,Y)$. Such learning set can then be used for computing the empirical risk, an estimate of the expected risk, that can be used for finding a good approximation of the optimal function $f^{*}$ that minimizes the aforementioned expectation. When it comes to TL, however, we assume that next to the information contained within $\mathcal{L}$, the learning algorithm also has access to an additional learning set defined as $\mathcal{L}^{\prime}$. Such a learning set can then be used alongside $\mathcal{L}$ for finding a function that better minimizes the loss function $\ell$. In this work, we consider $\mathcal{L}^{\prime}$ to be the ImageNet dataset, and $f$ to come in the form of either a pre-trained Convolutional Neural Networks or a pre-trained Vision Transformers, two types of neural networks that over the years have demonstrated exceptional abilities in tackling problems modeled by high dimensional and spatially organized inputs such as images, videos, and text.


\subsection{Related Works}
\label{sec:related_work}
While to this date, countless examples have studied the TL potential of CNNs \cite{Mormont_2018_CVPR_Workshops,vandaele2021deep,talo2019application,minaee2020deep}, the same cannot yet be said for VTs. In fact, papers that have so far investigated their generalization properties are much rarer. Yet, some works have attempted to compare the TL potential of VTs to that of CNNs, albeit strictly outside the artistic domain. E.g., in \cite{matsoukas2021time} the authors consider the domain of medical imaging and show that if CNNs are trained from scratch, then these models outperform VTs; however, if either an off-the-shelf feature extraction approach is used, or a self-supervised learning training strategy is followed, then VTs significantly outperform their CNNs counterparts. Similar results were also observed in \cite{zhou2021convnets} where the authors show that both in a single-task learning setting and in a multi-task learning one, transformer-based backbones outperformed regular CNNs on 13 tasks out of 15. Positive TL results were also observed in \cite{li2021benchmarking}, where the TL potential of VTs is studied for object detection tasks, in \cite{xue2021transfer} where the task of facial expression recognition is considered, and in \cite{duong2021detection}, where similarly to \cite{matsoukas2021time}, the authors successfully transfer VTs to the medical imaging domain. The generalization properties of VTs have also been studied outside of the supervised learning framework: E.g. in \cite{caron2021emerging} and \cite{chen2021empirical}, a self-supervised learning setting is considered. In \cite{caron2021emerging} the authors show that VTs can learn features that are particularly general and well suited for TL, which is a result that is confirmed in \cite{chen2021empirical}, where the authors show that VTs pre-trained in a self-supervised learning manner transfer better than the ones trained in a pure supervised learning fashion. While all these works are certainly valuable, it is worth noting that, except for \cite{matsoukas2021time} and \cite{duong2021detection}, all other studies have performed TL strictly within the domain of natural images. Therefore, the following research question 
\begin{center}
\textit{``How transferable are the representations of pre-trained VTs when it comes to the non-natural image domain?''}     
\end{center}
remains open. We will now present our work that helps us answer this question by considering \textbf{image datasets from the artistic domain.} Three art classification tasks are described, which serve as a surrogate for identifying which family of pre-trained networks, among CNNs and VTs, transfers better to the non-natural realm.

\section{Methods}
\label{sec:methods}
Our experimental setup is primarily inspired by the work presented in Sabatelli et al. \cite{sabatelli2018deep}, where the authors report a thorough empirical analysis that studies the TL properties of pre-trained CNNs that are transferred to different art collections. More specifically, the authors investigate whether popular neural architectures such as \texttt{VGG19} \cite{simonyan2014very} and \texttt{ResNet50} \cite{he2016deep}, which come as pre-trained on the ImageNet1k \cite{deng2009imagenet} dataset, can get successfully transferred to the artistic domain. The authors consider three different classification problems and two different TL approaches: an off-the-shelf (OTS) feature extraction approach where pre-trained models are used as pure feature extractors and a fine-tuning approach (FT) where the pre-trained networks are allowed to adapt all of their pre-trained parameters throughout the training process. Their study suggests that all the considered CNNs can successfully get transferred to the artistic domain and that a fine-tuning training strategy yields substantially better performance than the OTS one. In this work we investigate whether these conclusions also hold for transformer-based architectures, and if so, whether this family of models outperforms that of CNNs. 

\subsection{Data} \label{methods:dataset}

\begin{figure}[tb]
    \centering
    \includegraphics[width=\textwidth]{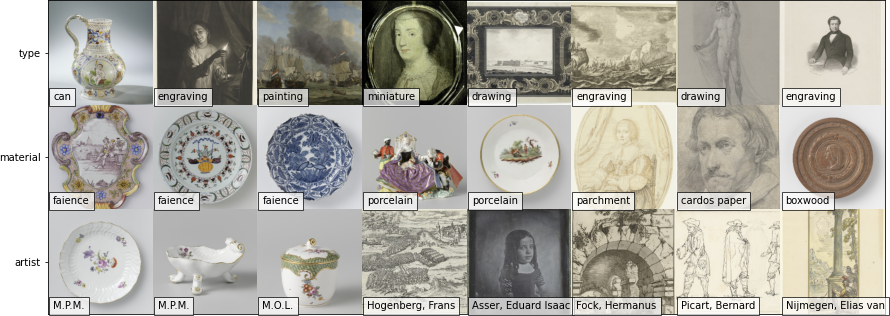}
    \caption{Samples from the three distinct classification tasks used throughout this paper.}
    \label{methods:img:examples}
\end{figure}

Similar to Sabatelli et al. \cite{sabatelli2018deep} we also use data stemming from the \textit{Rijks\-mu\-seum Challenge} dataset \cite{mensink14icmr} (see Fig. \ref{methods:img:examples} for an illustration).
This dataset consists of a large collection of digitized artworks that come together with \texttt{xml}-formatted metadata which can be used for defining a set of supervised learning problems \cite{mensink14icmr,strezoski2017omniart}. Following \cite{mensink14icmr} and \cite{sabatelli2018deep} we focus on three, well-known, classification problems namely: (1) \textit{Type classification}, where the goal is to train a model such that it is able to distinguish classes such as `painting', `sculpture', `drawing', etc.; (2) \textit{Material classification}, where the classification problem is defined by labels such as `paper', `porcelain', `silver', etc.; and finally, (3) \textit{Artist classification}, where, naturally, the model has to predict who the creator of a specific artwork is. While the full dataset contains 112,039 images, due to computational reasons in the present study we only use a fraction of it as this allows us to run shorter yet more thorough experiments. A smaller dataset also allows us to study an additional research question which we find worth exploring and that \cite{sabatelli2018deep} did not consider in their study, namely: \textit{``How well do CNNs (and VTs) transfer when the size of the artistic collection is small?''} To this end we decided to select the 30 most occurring classes within their dataset and to set a cap of 1000 randomly sampled instances per class. Table \ref{methods:datasets} summarizes the datasets used in the present study. Between brackets we report values as they were before balancing operations were performed.
For all of our experiments we use $5$ fold cross-validation and use 80\% of the dataset as training-set, 10\% as validation-set and 10\% as testing-set. 



\begin{table}[tb]
\centering
\caption{Overview of the used datasets. Values between brackets show the situation before balancing operations were performed. `Sample overlap' gives the average overlap between 2 of the 5 randomly generated sets per task ($i$ and $j$ where $i \neq j$).}
\small
\begin{tabular}{llll}
\hline
\textbf{Task} & \textbf{\# Samples} & \textbf{\# Classes} & \textbf{Sample overlap} \\ \hline
Type classification & 9607 (77628) & 30 (801) & 0.686 \\
Material classification & 7788 (96583) & 30 (136) & 0.798 \\
Artist classification & 6530 (38296) & 30 (8592) & 1 \\ \hline
\end{tabular}
\label{methods:datasets}
\end{table}


\subsection{Neural Architectures}  \label{methods:models}
In total, we consider eight different neural architectures, of which four are CNN-based networks while the remaining four are VT-based models. All models are pre-trained on the ImageNet1K dataset. When it comes to CNNs we consider \texttt{ResNet50} \cite{he2016deep} and \texttt{VGG19} \cite{simonyan2014very}, as the first one has widely been adopted by researchers working at the intersection of computer vision and digital heritage \cite{zhong2020fine,bai2021explain}, whereas the latter was among the best performing models considered by Sabatelli et al. \cite{sabatelli2018deep}. Next to these two architectures we also consider two additional, arguably more recent networks, namely \texttt{ConvNext} \cite{liu2022convnet}, which is a purely CNN-based model that is inspired by VTs' recent successes, and that therefore fits well within the scope of this work, and \texttt{EfficientNetV2} \cite{tan2021efficientnetv2}, which is a network that is well known for its computational efficiency and potentially faster training times. Regarding the VTs, we use models that have $16\times16$ patch sizes. As a result, we start by considering the first original VT model presented in \cite{dosovitskiy2020image} which we refer to as \texttt{ViT}. We then consider the \texttt{Swin} architecture \cite{liu2021swin} as it showed promising TL performance in \cite{zhou2021convnets} and the \texttt{BeiT} \cite{bao2022beit} and the \texttt{DeiT} \cite{touvron2021training} transformers. The usage of \texttt{DeiT} is motivated by Matsoukas et al. work \cite{matsoukas2021time} reviewed in Section \ref{sec:related_work} where the authors compared it to \texttt{ResNet50}. Yet, note that in this work, we use the version of the model known as the `base' version that does not take advantage of distillation learning in the FT stage. All VT-based models have $\approx 86$ million trainable parameters, and so does \texttt{ConvNext}. \texttt{ResNet50} and \texttt{EfficientNetV2} are, however much smaller networks as they come with $\approx25.6$ and $\approx13.6$ million trainable parameters, respectively. Lastly, \texttt{VGG19} is by far the largest model of all with its 143.7 million trainable parameters.


\subsection{Training Procedure}
For all experiments, images are resized to a $224 \times 224$ resolution by first scaling them to the desired size along the shortest axis (retaining aspect ratio), and then taking a center crop along the longer axis. In addition, all images are normalized to the RGB mean and standard deviation of the images presented within the ImageNet1K dataset ([0.485, 0.456, 0.406] and [0.229, 0.224, 0.225]). 
For all models we replace the final linear classification layer with a new layer with as many output nodes as there are classes to classify ($C$) within the dataset and optimize the parameters of the model $\theta$ such that the categorical cross-entropy loss function 
\begin{equation}
	\mathscr{L}(\theta) = - \mathds{E}_{(\mathbf{x},y)\sim P(X,Y)} \sum_{i=1}^{C} 1(y=i) \log p_\text{model} f_i(\mathbf{x};\theta),
	\label{eq:cross_entropy}
\end{equation}
is minimized.

Training is regularized through the early stopping method, which interrupts training if for 10 epochs in a row no improvement on the validation loss is observed. The model with the lowest validation loss is then benchmarked on the final testing set. For our OTS experiments, we use the Adam optimizer \cite{kingma2014adam} with standard \texttt{PyTorch} \cite{paszke2017automatic} parameters (\texttt{lr}=1e-3, $\beta_1=0.9$, $\beta_2=0.999$), and use a batch size of 256. For the FT experiments, hyperparameters are partially inspired by \cite{matsoukas2021time,zhou2021convnets}. We again use the Adam optimizer, this time initialized with \texttt{lr=\text{1e-4}} which gets reduced by a factor of 10 after three epochs without improvements on the validation loss; inspired by \cite{masters2018revisiting}, the batch size is now reduced to 32; whereas label smoothing (0.1) and dropout (p=0.2) are used for regularizing training even further. Finally, input images are augmented with random horizontal flips and rotations in a $\pm 10 ^\circ$ range.

\subsection{Hardware and Software}
All experiments are conducted on a single compute node containing one 32 GB \texttt{Nvidia V100 GPU}. The FT experiments take advantage of the V100's mixed precision acceleration. An exception is made for the \textit{Type Classification} experiment, as this one is also used to compare OTS TL and FT in terms of time/accuracy trade-offs (see Sec. \ref{sec:additional_studies} for further details). The \texttt{PyTorch} machine learning framework \cite{paszke2017automatic} is used for all experiments, and many pre-trained models are taken from its \texttt{Torchvision} library. Exceptions are made for \texttt{EfficientNetV2}, \texttt{Swin}, \texttt{DeiT} and \texttt{Beit}, which are taken from the \texttt{Timm} library\footnote{\url{https://timm.fast.ai/}.}. We release all source code and data on the following GitHub link\footnote{\url{https://github.com/IndoorAdventurer/ViTTransferLearningForArtClassification}}.



\section{Results}
\label{sec:results}

We now present the main findings of our study and report results for the three aforementioned classification problems and for all previously introduced architectures. All networks are either trained with an OTS training scheme (Sec. \ref{results:ots}) or with a FT one (Sec. \ref{results:ft}). Results come in the form of line plots and tables: the former visualize the performance of all models in terms of accuracy on the validation set, whereas the latter report the final performance that the best validation models obtained on the separate testing-set. All line plots report the average accuracy obtained across five different experiments, while the shaded areas correspond to the standard error of the mean ($\pm s \div \sqrt{N}$, where $N=5$). The dashed lines represent CNNs, whereas continuous lines depict VTs. Plots end when an early stop occurred for the first of the five trials. Regarding the tables, a green-shaded cell marks the best overall performing model, while yellow and red cells depict the second best and worst performing networks, respectively. We quantitatively assess the performance of the models with two separate metrics: the accuracy and the balanced accuracy, where the latter is defined as the average recall over all classes. Note that compared to the plain accuracy metric, the balanced accuracy allows us to penalize type-II errors more when it comes to the less occurring classes within the dataset.

\subsection{Off-The-Shelf Learning} \label{results:ots}

\begin{figure}[tbh]
    \centering
    \def\svgwidth{\textwidth}
    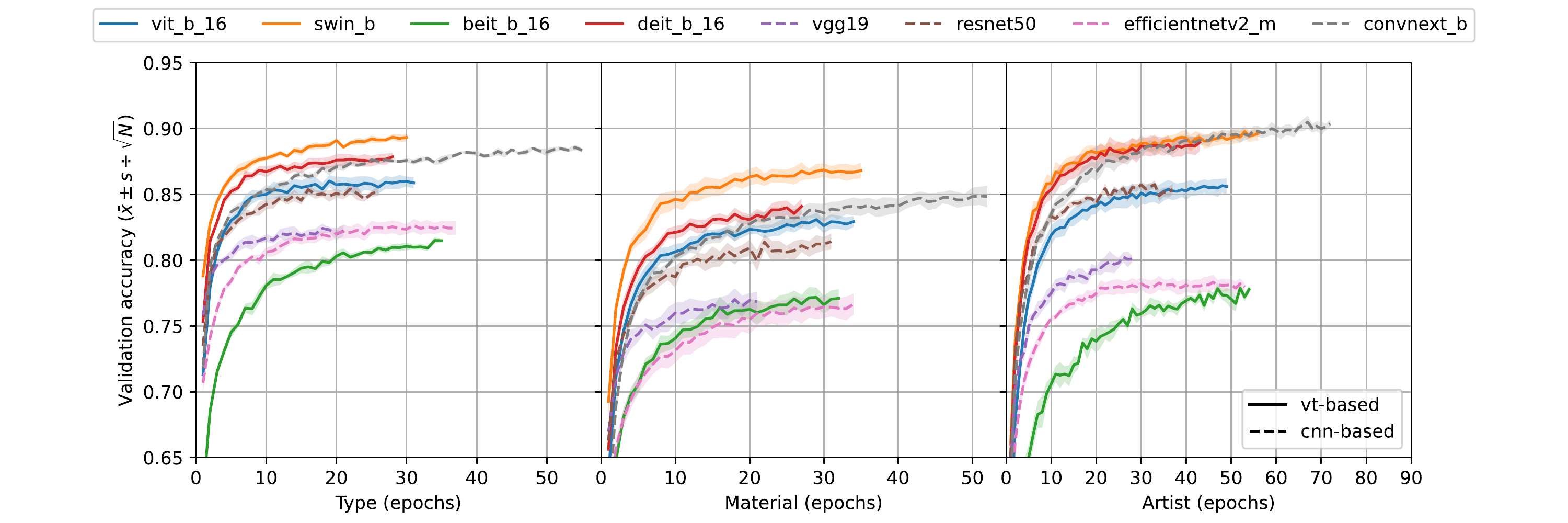
    \caption{The validation accuracy obtained by all architectures when trained with an off-the-shelf (OTS) feature extraction approach. We can see that the best performing models are the VT \texttt{Swin} and the CNN \texttt{ConvNext}.}
    \label{results:img:ots}
\end{figure}

\begin{figure}[tbh]
    \centering
    \def\svgwidth{\textwidth}
    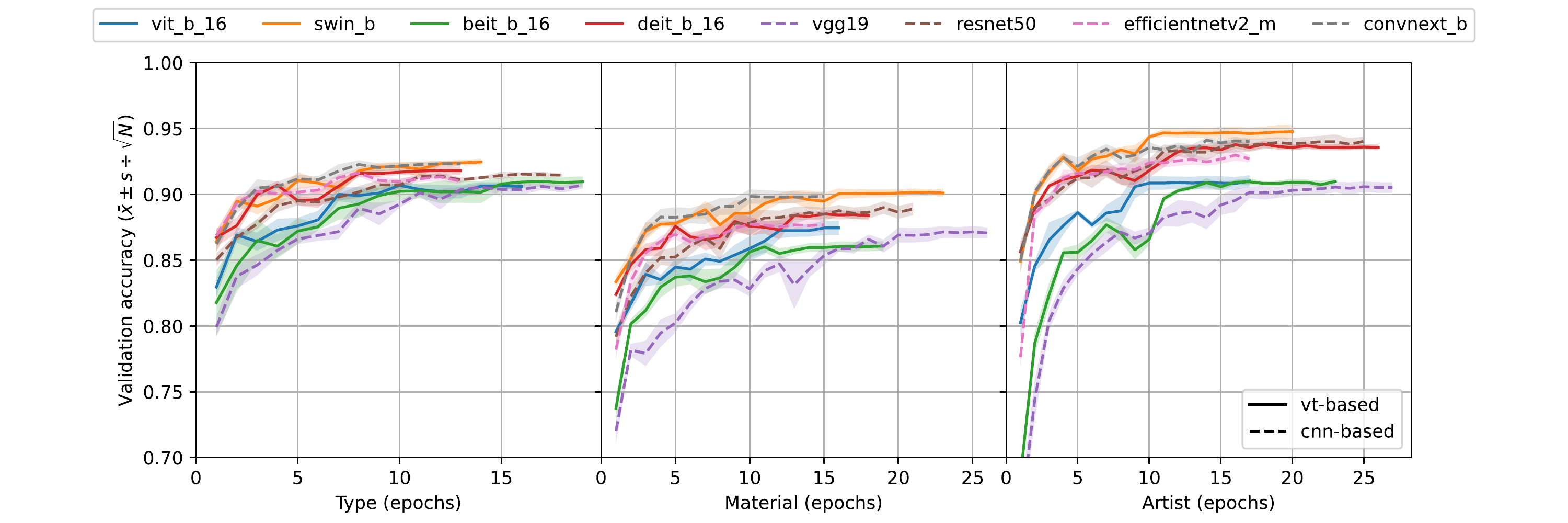
    \caption{The validation accuracy learning curves obtained by all models when a fine-tuning FT training approach is followed. We can again see that the \texttt{Swin} and \texttt{ConvNext} architectures are among the best performing networks.}
    \label{results:img:ft}
\end{figure}

For this set of experiments, we start by noting that overall, both the CNNs and the VTs can perform relatively well on all three different classification tasks. When it comes to (a) \textit{Type Classification} we can see that all models achieve a final accuracy between $\approx80\%$ and $\approx90\%$, whereas on (b) \textit{Material Classification} the performance deviates between $\approx 75\%$ and $85\%$, and on (c) \textit{Artist Classification} we report accuracies between $75\%$ and $90\%$. Overall, however, as highlighted by the green cells in Table \ref{results:tab:ots}, the best performing model on all classification tasks is the VT \texttt{Swin}, which confirms the good TL potential that this architecture has and that was already observed in \cite{zhou2021convnets}. Yet, we can also observe that the second-best performing model is not a VT, but the \texttt{ConvNext} CNN. Despite performing almost equally well, it is important to mention though that \texttt{ConvNext} required more training epochs to converge when compared to \texttt{Swin}, as can clearly be seen in all three plots reported in Fig. \ref{results:img:ots}. We also note that on the \textit{Type Classification} task the worst performing model is the \texttt{Beit} transformer ($82.26\%$), but when it comes to the classification of the materials and artists the worst performing model becomes \texttt{EfficientNetV2} with final accuracies of $75.96\%$ and $73.92\%$ respectively. Among the different VTs the \texttt{Beit} transformer also appears to be the architecture that requires the longest training. In fact, as can be seen in Fig. \ref{results:img:ft}, this network does not exhibit substantial ``Jumpstart-Improvements'' as the other VTs (learning starts much lower in all plots as is depicted by the green lines).

Several other interesting conclusions can be made from this experiment: we observed that the \texttt{VGG19} network yielded worse performance than \texttt{ResNet50}, a result which is not in line with what was observed by Sabatelli et al. \cite{sabatelli2018deep} where \texttt{VGG19} was the best performing network when used with an OTS training scheme. Also, differently from \cite{sabatelli2018deep}, the most challenging classification task in our experiments appeared to be that of \textit{Material Classification} as it resulted in the overall lowest accuracies. These results seem to suggest that even though both studies considered images stemming from the same artistic collection, the training dataset size can significantly affect the TL performance of the different models, a result which is in line with \cite{sabatelli2022contributions}.

In general, VTs seem to be very well suited for an OTS transfer learning approach, especially regarding the \texttt{Swin} and \texttt{DeiT} architectures which performed well across all tasks. Equally interesting is the performance obtained by \texttt{ConvNext} which is by far the most promising CNN-based architecture. Yet, on average, the performance of the VTs is higher than the one obtained by the CNNs: on \textit{Type Classification} the first ones perform on average $\approx 86.5$\% while the CNNs reach an average classification rate of $\approx 85.51\%$, whereas on \textit{Material} and \textit{Artist Classification} VTs achieve average accuracies of $\approx81.8\%$ and $\approx85.8\%$ respectively, whereas CNNs $\approx79.5\%$ $\approx81.2\%$. These results suggest that this family of methods is better suited for OTS TL when it comes to art classification problems.

\begin{table}[tb]
\caption{The performance of all models trained with an OTS approach on the final testing sets. We can see that the best overall model is the \texttt{Swin} transformer (green cells) followed by the \texttt{ConvNext} CNN architecture (yellow cells). The worst performing models are \texttt{Beit} when it comes to type classification and \texttt{EfficientNetV2} when it comes to material and artist classification.}
\centering
\resizebox{\columnwidth}{!}{%
\begin{tabular}{lllllll}
\hline
\textbf{Model} & \textbf{Type} &  & \textbf{Material} & & \textbf{Artist} & \\
& \textbf{Accuracy} & \textbf{Bal. accuracy} & \textbf{Accuracy} & \textbf{Bal. accuracy} & \textbf{Accuracy} & \textbf{Bal. accuracy} \\ \hline
\textbf{vit\_b\_16} & 86.06\% $_{\pm 1.06\%}$ & 84.13\% $_{\pm 1.57\%}$ & 81.78\% $_{\pm 0.48\%}$ & 67.38\% $_{\pm 1.37\%}$ & 84.89\% $_{\pm 0.46\%}$ & 81.42\% $_{\pm 0.42\%}$  \\
\textbf{swin\_b} & \cellcolor{bestcol}89.43\% $_{\pm 0.93\%}$ & \cellcolor{bestcol}87.47\% $_{\pm 1.02\%}$ & \cellcolor{bestcol}85.87\% $_{\pm 0.35\%}$ & \cellcolor{bestcol}71.19\% $_{\pm 1.60\%}$ & \cellcolor{bestcol}90.40\% $_{\pm 0.65\%}$ & \cellcolor{bestcol}88.64\% $_{\pm 0.78\%}$  \\
\textbf{beit\_b\_16} & \cellcolor{worstcol}82.26\% $_{\pm 0.72\%}$ & \cellcolor{worstcol}77.75\% $_{\pm 0.27\%}$ & 76.87\% $_{\pm 0.96\%}$ & 60.16\% $_{\pm 1.56\%}$ & 79.70\% $_{\pm 0.69\%}$ & 75.35\% $_{\pm 1.09\%}$  \\
\textbf{deit\_b\_16} & 88.18\% $_{\pm 0.66\%}$ & 85.36\% $_{\pm 0.54\%}$ & 82.80\% $_{\pm 1.12\%}$ & 66.46\% $_{\pm 1.03\%}$ & 88.13\% $_{\pm 0.76\%}$ & 85.62\% $_{\pm 0.87\%}$  \\\hdashline
\textbf{vgg19} & 83.93\% $_{\pm 0.72\%}$ & 83.35\% $_{\pm 0.81\%}$ & 76.87\% $_{\pm 0.44\%}$ & 61.39\% $_{\pm 1.47\%}$ & 82.01\% $_{\pm 0.66\%}$ & 78.10\% $_{\pm 0.77\%}$  \\
\textbf{resnet50} & 85.51\% $_{\pm 0.64\%}$ & 82.33\% $_{\pm 1.85\%}$ & 80.99\% $_{\pm 0.82\%}$ & 65.51\% $_{\pm 0.93\%}$ & 87.71\% $_{\pm 1.06\%}$ & 85.12\% $_{\pm 1.34\%}$  \\
\textbf{eff. netv2\_m} & 83.41\% $_{\pm 0.76\%}$ & 82.05\% $_{\pm 1.25\%}$  & \cellcolor{worstcol}75.96\% $_{\pm 1.24\%}$ & \cellcolor{worstcol}59.15\% $_{\pm 1.24\%}$ & \cellcolor{worstcol}78.62\% $_{\pm 1.07\%}$ & \cellcolor{worstcol}73.92\% $_{\pm 0.96\%}$  \\
\textbf{convnext\_b} & \cellcolor{secondbestcol}89.19\% $_{\pm 0.64\%}$ & \cellcolor{secondbestcol}86.95\% $_{\pm 1.38\%}$ & \cellcolor{secondbestcol}84.14\% $_{\pm 0.92\%}$ & \cellcolor{secondbestcol}69.10\% $_{\pm 1.05\%}$ & \cellcolor{secondbestcol}90.13\% $_{\pm 0.94\%}$ & \cellcolor{secondbestcol}87.84\% $_{\pm 1.07\%}$  \\\hline
\end{tabular}
}
\label{results:tab:ots}
\end{table}

\subsection{Fine-Tuning} \label{results:ft}
When it comes to the fine-tuning experiments, we observe, in part, consistent results with what we have presented in the previous section. We can again note that the lower classification rates have been obtained when classifying the material of the different artworks, while the best performance is again achieved when classifying the artists of the heritage objects. The \texttt{Swin} VT remains the network that overall performs best, while the \texttt{ConvNext} CNN remains the overall second best performing model, even though on \textit{Artist Classification} it is slightly outperformed by \texttt{ResNet50}. Unlike our OTS experiments, however, this time, we note that VTs are on average the worst-performing networks.
When it comes to the \textit{Type Classification} problem, the \texttt{ViT} model achieves the lowest accuracy (although note that if the balance accuracy metric is used, then the worst performing network becomes \texttt{VGG19}). Further, when considering the classification of materials and artists, the lowest accuracies are instead achieved by the \texttt{Beit} model. While it is true that overall the performance of both CNNs and VTs significantly improves through fine-tuning, it is also true that, perhaps surprisingly, such a training approach reduces the differences in terms of performance between these two families of networks. This result was not observed in \cite{zhou2021convnets}, where VTs outperformed CNNs even in a fine-tuning training regime.
While these results indicate that an FT training strategy does not seem to favor either VTs or CNNs, it is still worth noting that this training approach is beneficial. In fact, no matter whether VTs or CNNs are considered, the worst-performing fine-tuned model still performs better than the best performing OTS network. The only exception to this is \textit{Material Classification}, where \texttt{BeiT} obtained a testing accuracy of 85.75\% after fine-tuning which is slighlty lower than the one obtained by the \texttt{Swin} model trained with an OTS training scheme (85.87\%).

\begin{table}[tb]
\caption{The testing set performance of all models trained with a FT approach. We can see that compared to the results presented in Table \ref{results:tab:ots}, all models perform substantially better, yet differences in terms of performance between the VTs and the CNNs now seem to be smaller.}
\centering
\resizebox{\columnwidth}{!}{%
\begin{tabular}{lllllll}
\hline
\textbf{Model} & \textbf{Type} &  & \textbf{Material} & & \textbf{Artist} & \\
& \textbf{Accuracy} & \textbf{Bal. accuracy} & \textbf{Accuracy} & \textbf{Bal. accuracy} & \textbf{Accuracy} & \textbf{Bal. accuracy} \\ \hline
\textbf{vit\_b\_16} & \cellcolor{worstcol}90.11\% $_{\pm 0.35\%}$ & 87.40\% $_{\pm 0.29\%}$ & 87.42\% $_{\pm 0.45\%}$ & 73.46\% $_{\pm 1.44\%}$ & 92.05\% $_{\pm 0.44\%}$ & 89.77\% $_{\pm 0.38\%}$  \\
\textbf{swin\_b} & \cellcolor{bestcol}92.17\% $_{\pm 0.98\%}$ & \cellcolor{secondbestcol}89.71\% $_{\pm 1.03\%}$ & \cellcolor{bestcol}89.35\% $_{\pm 0.68\%}$ & 77.16\% $_{\pm 2.98\%}$ & \cellcolor{bestcol}95.05\% $_{\pm 0.47\%}$ & \cellcolor{bestcol}93.94\% $_{\pm 0.81\%}$  \\
\textbf{beit\_b\_16} & 90.81\% $_{\pm 0.41\%}$ & 87.95\% $_{\pm 0.69\%}$ & \cellcolor{worstcol}85.74\% $_{\pm 0.37\%}$ & \cellcolor{worstcol}72.12\% $_{\pm 1.38\%}$ & \cellcolor{worstcol}91.27\% $_{\pm 1.13\%}$ & \cellcolor{worstcol}88.83\% $_{\pm 1.63\%}$  \\
\textbf{deit\_b\_16} & 91.78\% $_{\pm 0.64\%}$ & 89.22\% $_{\pm 0.90\%}$ & 87.85\% $_{\pm 1.12\%}$ & 74.42\% $_{\pm 1.99\%}$ & 93.37\% $_{\pm 1.15\%}$ & 91.67\% $_{\pm 1.55\%}$  \\\hdashline
\textbf{vgg19} & 90.54\% $_{\pm 0.37\%}$ & \cellcolor{worstcol}87.05\% $_{\pm 1.03\%}$ & 85.74\% $_{\pm 1.40\%}$ & 72.43\% $_{\pm 3.03\%}$ & 92.20\% $_{\pm 0.49\%}$ & 90.18\% $_{\pm 0.72\%}$  \\
\textbf{resnet50} & 91.78\% $_{\pm 0.44\%}$ & 88.24\% $_{\pm 0.59\%}$ & 88.69\% $_{\pm 0.99\%}$ & \cellcolor{secondbestcol}77.97\% $_{\pm 2.25\%}$ & \cellcolor{secondbestcol}94.72\% $_{\pm 0.74\%}$ & \cellcolor{secondbestcol}93.41\% $_{\pm 1.05\%}$  \\
\textbf{eff. netv2\_m} & 90.87\% $_{\pm 0.67\%}$ & 88.34\% $_{\pm 1.37\%}$ & 87.55\% $_{\pm 1.15\%}$ & 75.31\% $_{\pm 1.60\%}$ & 92.65\% $_{\pm 0.54\%}$ & 90.84\% $_{\pm 0.51\%}$  \\
\textbf{convnext\_b} & \cellcolor{secondbestcol}92.15\% $_{\pm 0.40\%}$ & \cellcolor{bestcol}89.82\% $_{\pm 1.18\%}$ & \cellcolor{secondbestcol}88.79\% $_{\pm 1.07\%}$ & \cellcolor{bestcol}78.40\% $_{\pm 1.26\%}$ & 94.60\% $_{\pm 0.54\%}$ & 93.13\% $_{\pm 0.61\%}$  \\\hline
\end{tabular}
}
\label{results:tab:ft}
\end{table}

\section{Discussion}
\label{sec:discussion}

Our results show that VTs possess strong transfer learning capabilities and that this family of models learns representations that can generalize to the artistic realm. Specifically, as demonstrated by our OTS experiments, when these architectures are used as pure feature extractors, their performance is, on average, substantially better than the one of CNNs. To the best of our knowledge, this is the first work that shows that this is the case for non-natural and artistic images. Next to be attractive to the computer vision community, as some light is shed on the generalization properties of VTs, we believe that these results are also of particular interest for practitioners working in the digital humanities with limited access to computing power. As the resources for pursuing an FT training approach might not always be available, it is interesting to know that between CNNs and VTs, the latter models are the ones that yield the best results in the OTS training regime. 

Equally interesting and novel are the results obtained through fine-tuning, where the performance gap between VTs and CNNs gets greatly reduced, with the latter models performing only slightly worse than the former. In line with the work presented in \cite{sabatelli2018deep}, which considered CNNs exclusively, we clearly show that this TL approach also substantially improves the performance of VTs. 

\section{Additional Studies}
\label{sec:additional_studies}

We now present four additional studies that we hope can help practitioners that work at the intersection of computer vision and the digital humanities. Specifically we aim to shed some further light into the classification performance of both CNNs and VTs, while also providing some practical insights that consider the training times of both families of models.

\subsection{Saliency Maps}

\begin{figure*}[tb]
     \includegraphics[width=\textwidth]{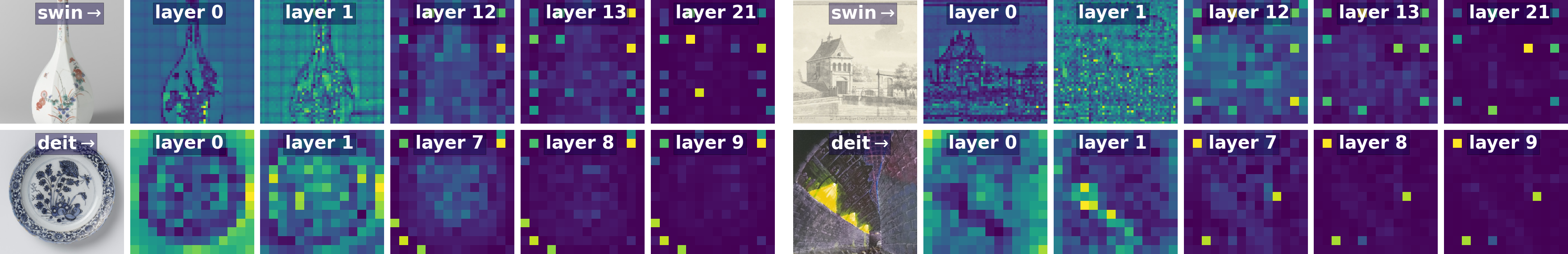}
     \caption{Saliency maps for different attention layers of successively deeper transformer blocks. The deeper the network, the lower level the representations learned by the Vision Transformer become.}
     \label{results:img:layers}
\end{figure*}

We start by performing a qualitative analysis that is based on the visual investigation of saliency maps. For this set of studies we consider the \texttt{ConvNext} CNN and the \texttt{Swin} and \texttt{DeiT} VTs. When it comes to the former, saliency maps are computed through the popular GradCam method \cite{selvaraju2017grad}, whereas attention-rollout is used when it comes to the \texttt{DeiT} architecture. \texttt{Swin}'s saliency maps are also computed with a method similar to GradCam, with the main difference being that instead of taking an average of the gradients per channel as weights, we directly multiply gradients by their activation and visualize patch means of this product. Motivated by the nice performance of VTs in the OTS transfer learning setting, we start by investigating how the representation of an image changes with respect to the depth of the network. While it is well known that CNNs build up a hierarchical representation of the input throughout the network, similar studies involving transformer-based architectures are rarer. In Fig. \ref{results:img:layers} we present some examples that show that also within VTs, the deeper the network becomes, the more the network starts focusing on lower level information within the image. 

\begin{figure}[htp]
\centering
\includegraphics[width=.45\textwidth]{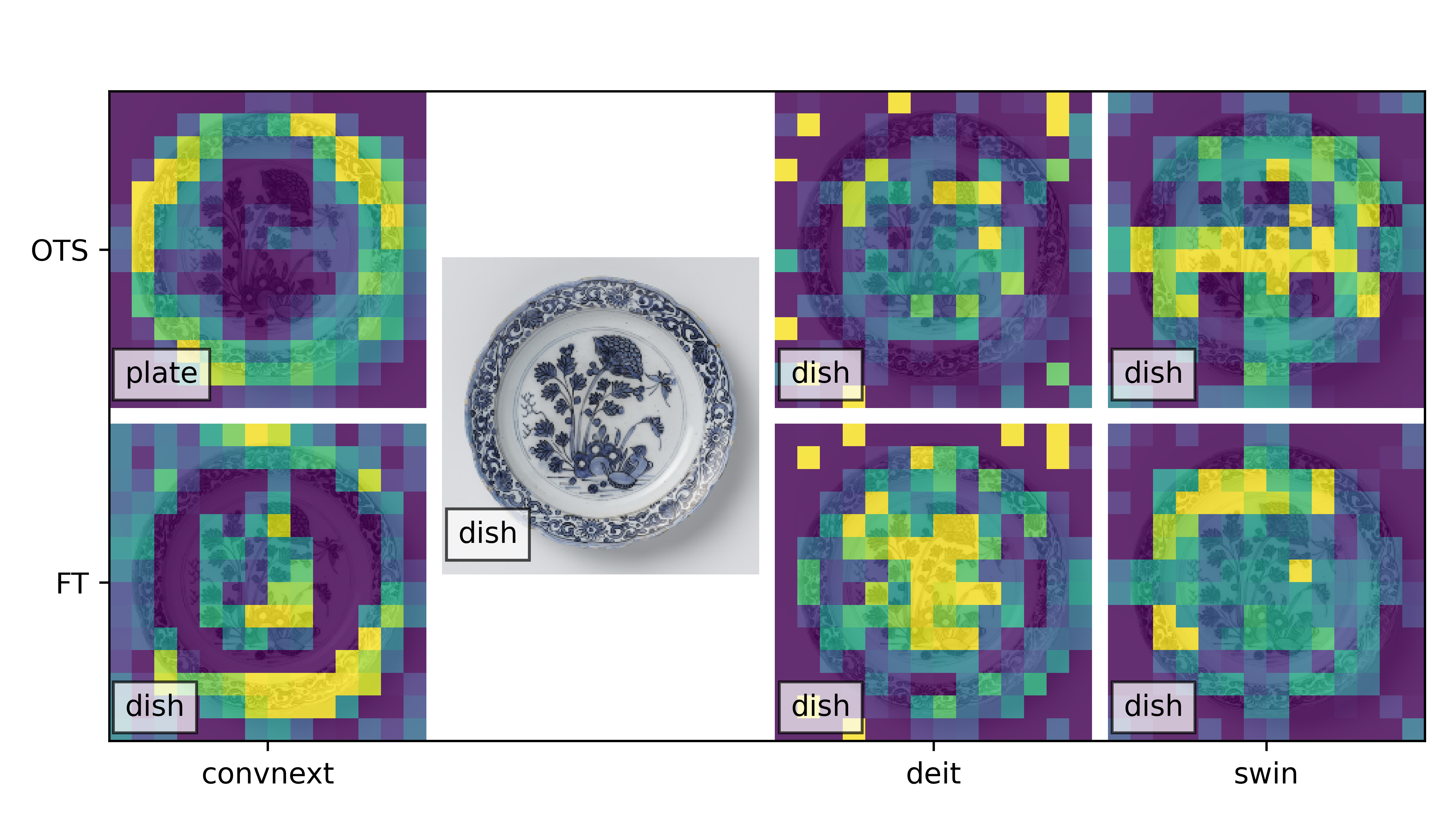}\hfill
\includegraphics[width=.45\textwidth]{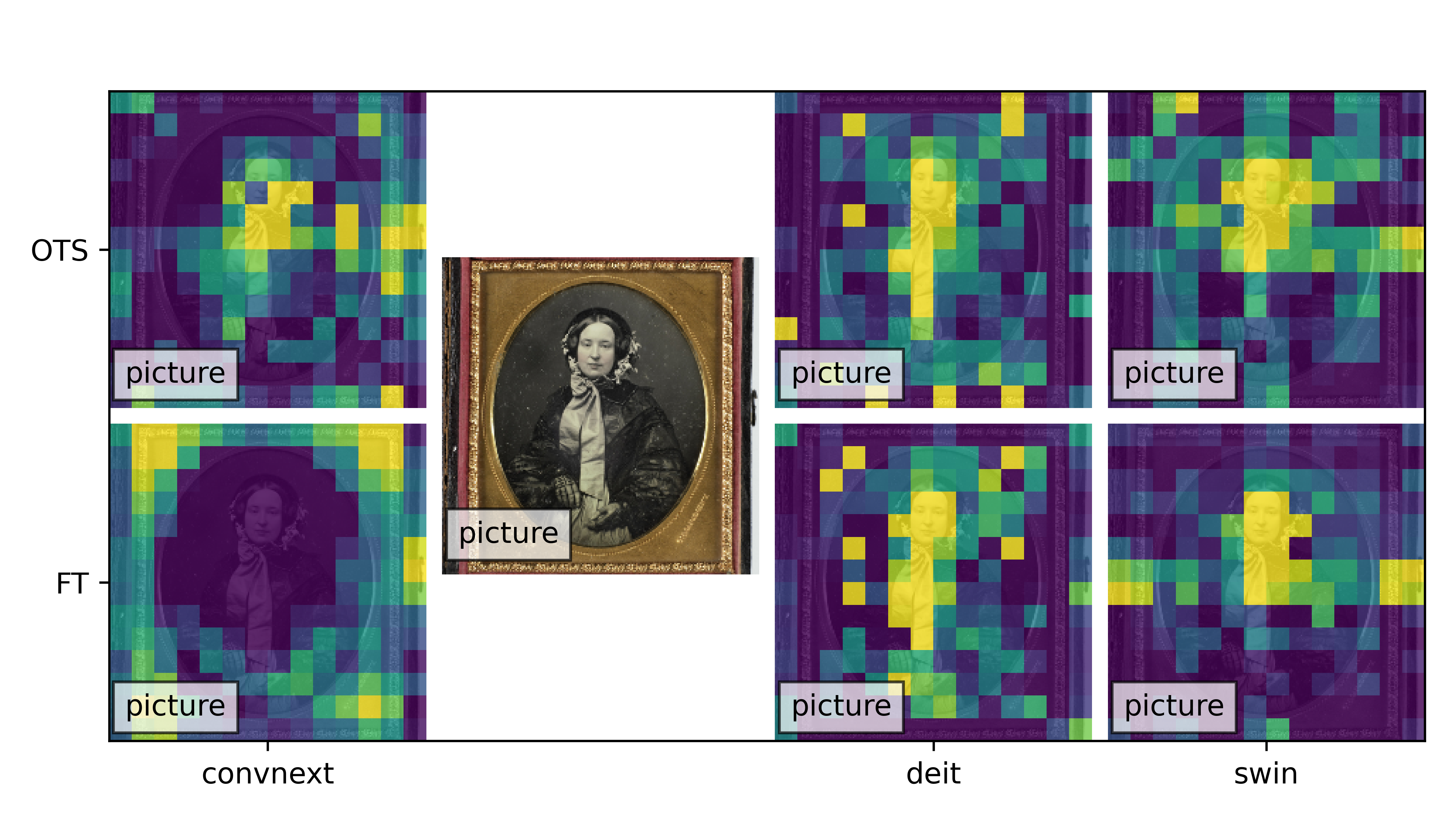}\hfill
\includegraphics[width=.45\textwidth]{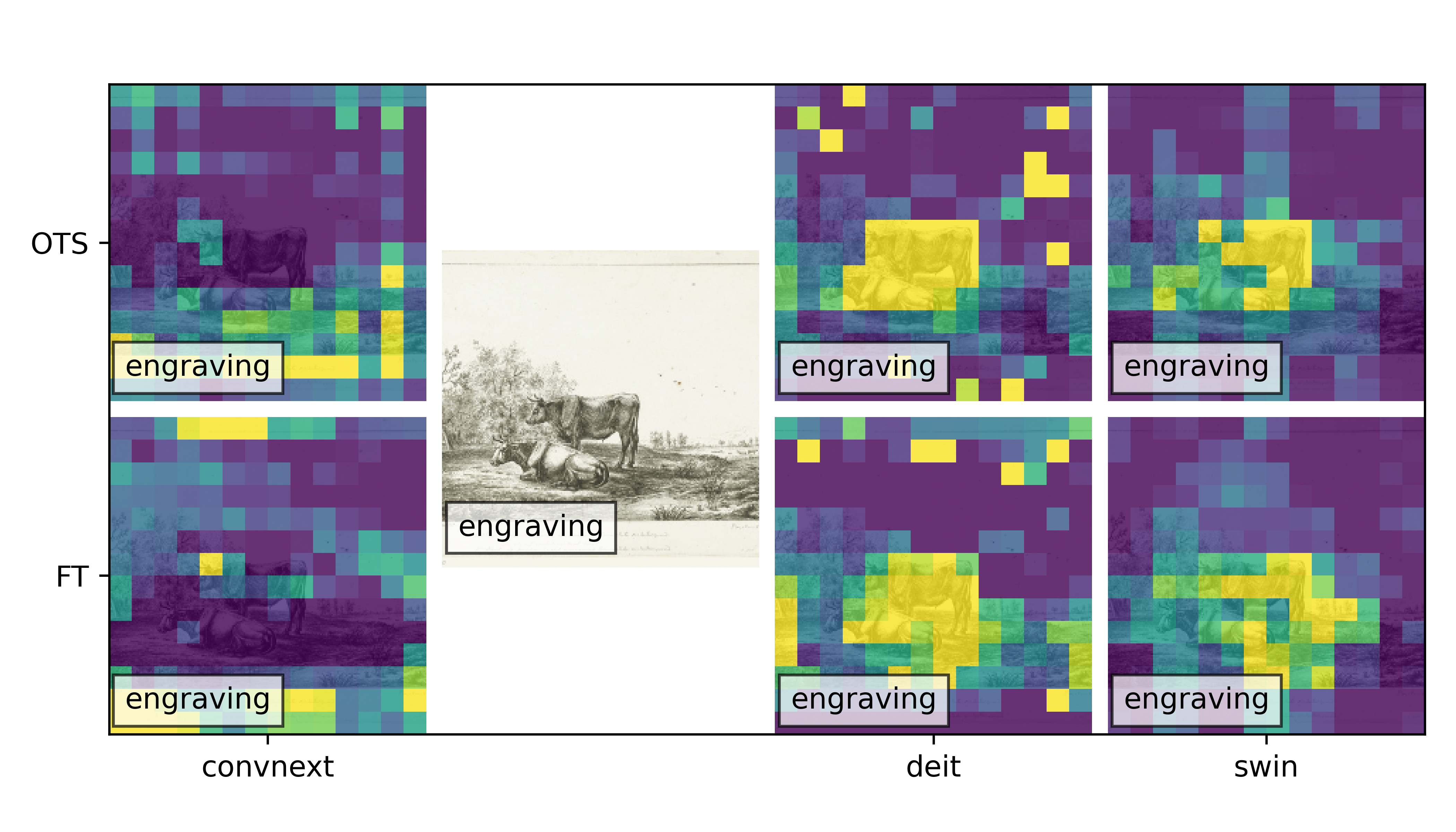}

\caption{Saliency maps computed for \texttt{ConvNext}, \texttt{DeiT} and \texttt{Swin} when classifying the \textit{``Type''} of three heritage objects. Saliency maps are computed both for the off-the-shelf experiments as well as for the fine-tuning ones.}
\label{fig:saliency_maps}

\end{figure}

In Fig. \ref{fig:saliency_maps} we show how different network architectures and TL approaches result in different saliency maps. In the first image of Fig. \ref{fig:saliency_maps} we can observe that the \texttt{ConvNext} architecture miss-classifies a `Dish' as a `Plate' when an OTS approach is used. Note that this is a mistake not being made by the transformer-based architectures. However, we observe that after the fine-tuning stage, the CNN can classify the image's \textit{Type} correctly after having shifted its attention toward the bottom and center of the dish rather than its top. 
Interestingly, none of the transformer-based architecture focuses on the same image regions, independently of whether an OTS or a FT approach is used. We can see that most saliency maps are clustered within the center of the image, both when an OTS training strategy is used and when an FT approach is adopted. When looking at the second image of Fig. \ref{fig:saliency_maps}, we see that all networks, independently from the adopted TL approach, correctly classify the image as a \textit{`Picture'}. Yet the saliency maps of the \texttt{ConvNext} change much more across TL approaches, and we again see that the transformer-based architectures focus more on the central regions of the image rather than on the borders. Similar behavior can also be observed in the last image of Fig. \ref{fig:saliency_maps}. 

\subsection{Dealing With Small Artistic Collections}
\begin{figure*}[tb]
    \centering
    \def\svgwidth{\textwidth}
    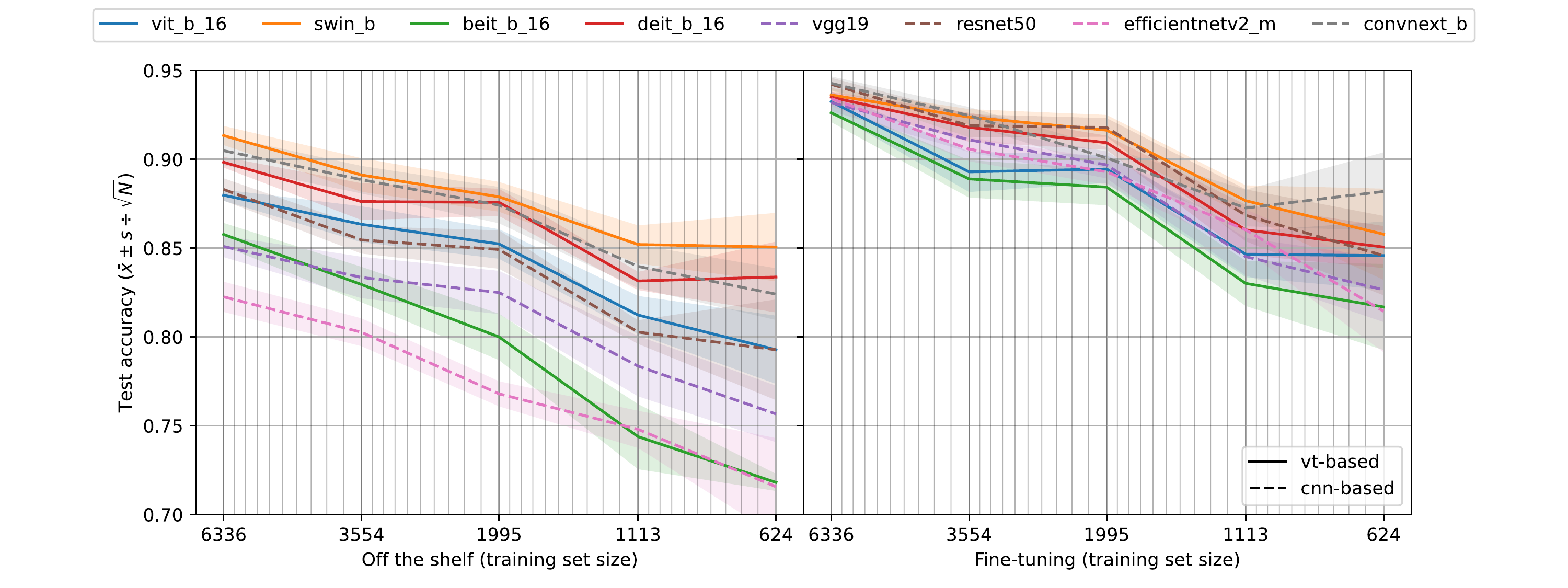
    \caption{The testing accuracy obtained on the \textit{Type Classification} problem when gradually reducing the size of the dataset. We can observe that different dataset sizes do not affect the best performing models, which as presented in Sec. \ref{sec:results} remain the \texttt{Swin} VT and the \texttt{ConvNext} CNN.}
    \label{results:img:scale}
\end{figure*}

It is not uncommon for heritage institutions to deal with datasets far smaller than those typically used by the computer vision community. While it is true that the number of samples used throughout this study is far smaller than the one used within the naturalistic domain, we still wondered whether the results reported in Sec. \ref{sec:results} would generalize to even smaller artistic collections. Inspired by \cite{NEURIPS2021_c81e155d} we have designed the following experiment where we consider the \textit{Type Classification} task. We consider the top 15 most occurring classes within the original dataset presented in Table \ref{methods:datasets} and scale the number of samples four times by a factor $\sqrt[4]{\frac{1}{10}} \approx 0.56$. This results in five datasets which are respectively 100\%, 56\%, 32\%, 18\% and 10\% the size of the original dataset shown in Table \ref{methods:datasets}. Note that we ensure that the distribution of samples per class remains the same across all datasets. We then trained all models following the exact experimental protocol described in Sec. \ref{sec:methods}. Our results are reported in Fig. \ref{results:img:scale} where we can observe how the testing accuracy decreases when smaller portions of the full dataset are taken. Note that the x-axes show a logarithmic scale, with the rightmost value being roughly 10\% the size of the leftmost one. For both an OTS approach (left plot of Fig. \ref{results:img:scale}), as well as a FT one (right plot of Fig. \ref{results:img:scale}), we show that the findings discussed in Sec. \ref{results:ots} and Sec. \ref{results:ft} generalize to smaller datasets.


\subsection{Training Times}

\begin{figure}{}
\centering
\def\svgwidth{5cm}
    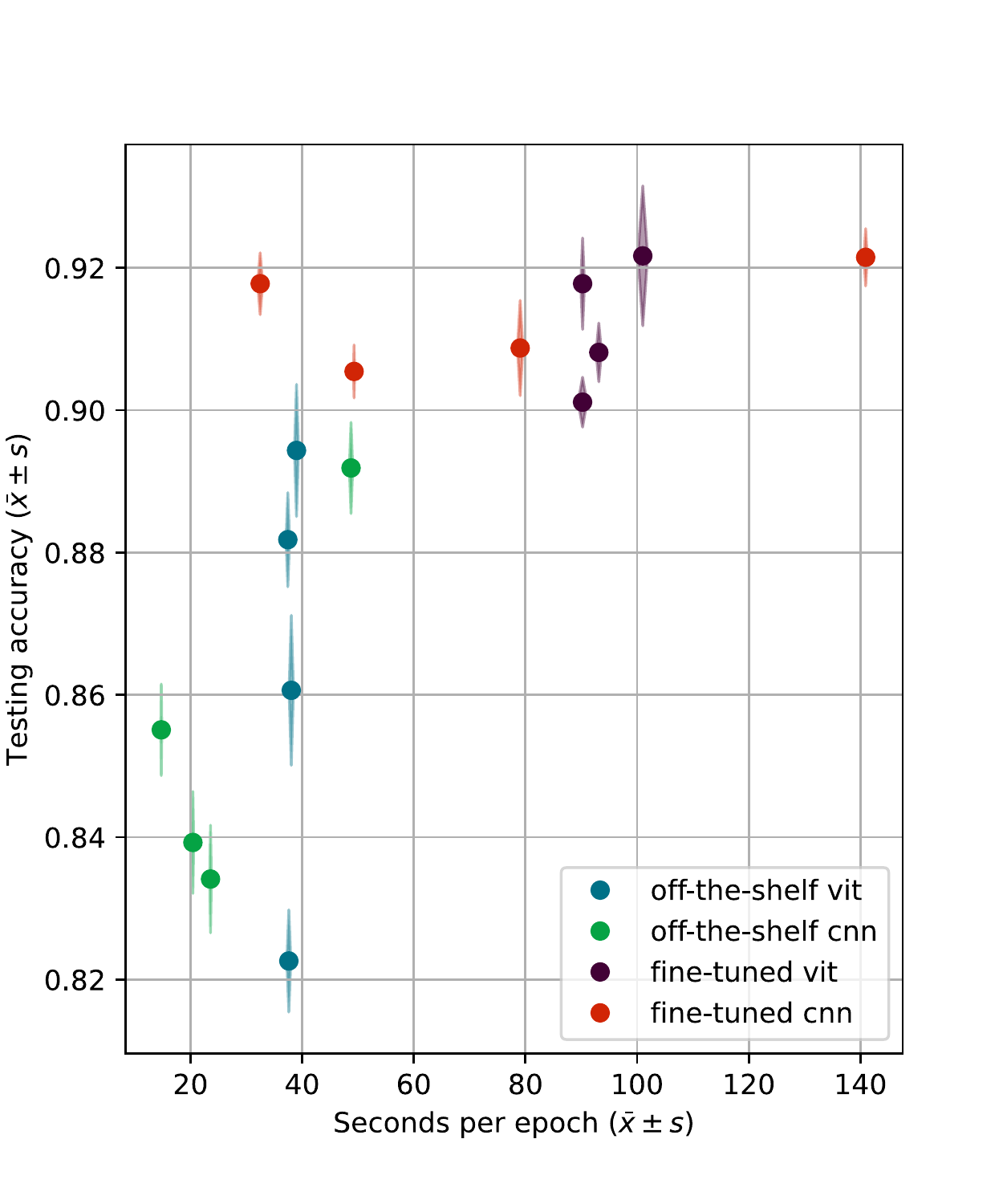
    \caption{Time/accuracy trade-offs for VTs and CNNs trained either in an off-the-shelf setting or with a fine-tuning strategy.}
    \label{results:img:ots_vs_ft_type}
\end{figure}

We now report some final results that describe the training times that are required by CNNs and VTs when trained with the aforementioned TL strategies. These results have also been obtained on the \textit{Type} classification task but with the V100s mixed precision capabilities disabled. In Fig. \ref{results:img:ots_vs_ft_type} we show the classification accuracy as reported in Table \ref{results:tab:ots} and \ref{results:tab:ft} and plot it against the average duration of one training epoch. This allows us to understand the time/accuracy trade-offs that characterize all neural networks. As shown by the blue dots, we can observe that when it comes to VTs trained in an OTS setting, all transformer-based architectures require approximately the same number of seconds to successfully go through one training epoch ($\approx 40$). This is, however, not true for the CNNs, which on average, require less time and for which there is a larger difference between the fastest OTS model (\texttt{ResNet50} requiring $\approx 20$ seconds) and the slowest model (\texttt{ConvNext}). While, as discussed in Sec. \ref{sec:results}, VTs are more powerful feature extractors than CNNs, it is worth noting that the gain in performance these models have to offer comes at a computational cost. Note, however, that this is not true anymore when it comes to an FT training regime, as all models (red and purple dots) now perform almost equally well. Yet it is interesting to point out that the computational costs required by the VTs stay approximately the same across all architectures, which is not the case for the CNN-based networks as there is a clear difference between the fastest fine-tuned network (\texttt{ResNet50}) and the slowest (\texttt{ConvNext}). We believe that the design of a transformer-based architecture that, if fine-tuned, results in the same training times as \texttt{ResNet50} provides an interesting avenue for future work.

\section{Conclusion}
This work examined how well VTs can transfer knowledge to the non-natural image domain. To this end, we compared four popular VT architectures with common CNNs, in terms of how well they transfer from ImageNet1k to classification tasks presented in the \textit{Rijksmuseum Challenge} dataset. We have shown that when fine-tuned VTs performed on par with CNNs and that they performed even better than their CNN counterparts when used as feature extractors. We believe that our study proves that VTs can become a valuable alternative to CNN-based architectures in the non-natural image domain and in the realm of artistic images specifically. Especially \texttt{Swin} and \texttt{DeiT} showed promising results throughout this study and, therefore, we aim to investigate their potential within the Digital Humanities in the future. To this end, inspired by \cite{strezoski2017omniart} we plan on using them in a multi-task learning setting; as backbone feature extractors when it comes to the object detection within artworks \cite{gonthier2018weakly,sabatelli2021advances}, and finally, in a self and semi-supervised learning setting. We also plan on performing a similar analysis for the \texttt{ConvNext} CNN, as this architecture showed very good performance as well.

To conclude, we believe that the study presented in this work opens the door for a more fundamental CV question that deserves attention: \textit{``What makes transformer-based architectures such powerful feature extractors?''}. We plan on investigating whether the results obtained within the domain of DH will also generalize to other non-natural image datasets with the hope of partially answering this question. 


\clearpage
%
%
\bibliographystyle{splncs04}
\bibliography{egbib}
\end{document}